%% file: anonymous-submission-latex-2026.tex
\documentclass[letterpaper]{article} 
\usepackage{aaai2026}
\nocopyright
\usepackage{times}  
\usepackage{helvet}  
\usepackage{courier}  
\usepackage[hyphens]{url}  
\usepackage{graphicx} 
\usepackage{natbib}  
\usepackage{caption} 
\usepackage{times}
\usepackage{soul}
\usepackage[utf8]{inputenc}
\usepackage{svg}

\usepackage{graphicx}
\usepackage{amsmath}
\usepackage{amsthm}
\usepackage{amssymb}
\usepackage{booktabs}
\usepackage{algorithm}
\usepackage{algorithmic}
\usepackage[switch]{lineno}
\usepackage{multirow}
\usepackage{color}
\usepackage{xcolor}
\input{math}
\usepackage{newfloat}
\usepackage{listings}
\DeclareCaptionStyle{ruled}{labelfont=normalfont,labelsep=colon,strut=off} 
\floatstyle{ruled}
\newfloat{listing}{tb}{lst}{}
\floatname{listing}{Listing}
\title{A Survey on Pre-Trained Diffusion Model Distillations}
\author{
    Xuhui Fan\textsuperscript{\rm 1}\\
    Zhangkai Wu\textsuperscript{\rm 1},
    Hongyu Wu\textsuperscript{\rm 1},\\
    Longbing Cao\textsuperscript{\rm 1}\\
}
\affiliations{
    \textsuperscript{\rm 1}School of Computing, Macquarie University, Macquarie Park, NSW 2109, Australia
}

\begin{document}

\maketitle

\begin{abstract}
Diffusion Models~(DMs) have become the dominant approach in Generative Artificial Intelligence (GenAI), owing to their remarkable performance in tasks such as text-to-image synthesis. However, practical DMs, such as stable diffusion, are typically trained on massive datasets and thus usually require large memory storage. At the same time, DMs usually require multiple steps, i.e. recursively evaluating the trained neural network, to generate a high-quality image, which results in significant computational costs for sample generation. As a result, distillation methods on pre-trained DMs have become popular choices to develop smaller-size and more efficient models capable of fast image generation. When various distillation methods have been developed from different perspectives, there is an urgent need for a systematic survey, particularly from a methodological perspective. This survey presents a systematic and up-to-date review on existing distillation methods. Particularly, these distillation methods are divided into three categories: fidelity distillation, trajectory distillation and adversarial distillation, in which each category is further grouped based on their detail focus. We discuss current challenges and outline future research directions as the conclusion of this survey. 
\end{abstract}

\section{Introduction}
\label{sec:introduction}
Generative Artificial Intelligence~(GenAI) has witnessed remarkable achievements in recent years~\cite{ho2020denoising,song2020denoising,wu2024evae,wu2025wavae,wu2023c}, with DMs emerging as dominant models. These models, such as stable diffusion~\cite{rombach2022high,esser2024scaling}, have demonstrated exceptional performance across a wide range of generative tasks, including text-to-image synthesis~\cite{rombach2022high,esser2024scaling,wu2024paramrel,wu2025sepdiff,wu2025progdiffusion}, text-to-audio generation~\cite{liu2023audioldm,wu2023large,evans2024fast}, and beyond~\cite{watson2023novo,yim2023se,wu2024protein}.

However, the multi-step sample generation mechanism in DMs makes it unappealing in practice, especially in low-resource environments. Unlike single-step generative models such as Generative Adversarial Networks~(GANs)~\cite{goodfellow2020generative}, DMs generate samples through an iterative process that requires to recursively evaluate the trained neural network function. This mechanism, while effective in producing high-quality outputs (e.g. images), incurs substantial computational costs for these high Number of Function Evaluations~(NFE). Furthermore, training practical DMs typically requires massive data, which challenges the training process. As a result, deploying DMs in real-world applications, where efficiency and speed are critical, remains a significant challenge.

To address these limitations, pre-trained diffusion distillation methods have gained attention as promising solutions. Distillation techniques aim to create smaller-size, more efficient models capable of generating high-quality samples in fewer steps, thereby reducing computational overhead. These methods vary widely in their approaches, ranging from fidelity loss function design to the denoising trajectory optimization. Despite the growing interest in distillation, the field lacks an up-to-date and systematic survey of these methods, particularly from a methodological perspective. 

In this paper, we aim to fulfill this gap by providing a thorough review of DMs distillation methods. We begin with a brief overview of DMs, highlighting their strengths and limitations. Then, we systematically examine distillation methods and categorize them into:
\begin{description}
    \item[Fidelity distillation] We will review distillation methods focusing on their outputs (e.g. images), including the image values, image distributions, one-step denoised images, and Fisher divergences of the image distributions. 
    \item[Trajectory distillation] It focuses on the trajectory between the noise distribution and the data distribution. Trajectory distillation includes consistency distillation, rectified flow distillation and their integration.
    \item[Adversarial distillation] It uses the adversarial loss as the distillation loss to optimize the generator.
\end{description}
Finally, we discuss the current challenges and propose potential future directions for research.

By synthesizing existing knowledge and offering new perspectives, this survey aims to serve as a valuable resource for researchers and practitioners in GenAI. Our goal is to not only summarize the state of the art but also inspire innovative solutions in further improving the distilling performance. 

\subsubsection{Difference to existing surveys} While there have been several works~\cite{yang2022diffusion,huang2024diffusion,shuai2024survey} discussing and reviewing DMs, their scope and focuses are different from those of this survey. For example,~\cite{yang2022diffusion} gives a general overview of DMs. Only one paragraph is used to discuss one basic method progressive distillation~\cite{salimans2022progressive}. \cite{huang2024diffusion,shuai2024survey} focus on using DMs for image editing and multi-modal generation. Comparing to these existing surveys, our survey focuses on {pre-trained DMs distillation}. 

The existing survey~\cite{luo2023comprehensive} might be closest to our survey, as it also discusses DMs distillation methods. However, this survey was written around two years ago, while important developments, especially consistency models, rectified flow distillation methods and adversarial distillation, are developed within the last two years. More importantly, our survey provides two new perspectives on fidelity distillation, trajectory distillation, and adversarial distillation, to summarize these methods. The blog~\cite{dieleman2024distillation} also discusses diffusion distillation methods. While it focuses on the intuition of each method, a systematic and more solid discussion is needed. In our survey, we use one notation system to discuss motivations, procedures and objective functions of each distillation methods such that they can be conveniently compared and precisely evaluated.

\subsubsection{Structure of the survey} The remainder of this survey is organized as follows. We first cover the preliminary knowledge of the foundational concepts in DMs. Then, we present an overview of DMs distillation methods, including the motivations for categorizing these methods. The next three sections introduce each categories of distillation methods in detail, including the fidelity distillation which is driven by the output difference objective, the trajectory distillation which focuses on the transition trajectory between random noise and clean data, and the adversarial distillation. This survey is concluded with discussion on current challenges and potential future directions in the field of pre-trained diffusion model distillation methods.

\section{Preliminary}
\label{sec:preliminaries}
\subsubsection{Notation Setup.} We consider DMs' forward diffusion process and backward denoising process to be within the time interval $[0, 1]$. The corresponding time steps are ordered as $1=t_n<t_{n-1}<\ldots<t_1=0$. Since most DMs have been applied in image synthesis tasks, we use images to denote the samples. This denotes $\mbx_0$ as the clean image, $\mbx_1$ as random noise from standard Gaussian distributions, and $\mbx_t$ as the noisy image at the time step $t$. 

In addition, we use $\mbtheta$ to represent the parameters of the pre-trained model. With an abuse of notation, we write the score function of pre-trained DMs  as $\mbepsilon_{\mbtheta}(\mbx_t, t)$, and write the velocity of the pre-trained rectified flow as $\mbv_{\mbtheta}(\mbx_t, t)$. We use $\mbphi$ to denote the parameters of the student model. {We let $\mbx_{t_{i+1}}=g_{\mbtheta}(\mbx_{t_i}, {t_i})$ and $\mbx_{t_{i+1}}=f_{\mbphi}(\mbx_{t_i}, {t_i})$ denote pre-trained model and distillation model's projected locations at time step $t_{i+1}$.}

\subsubsection{Diffusion models}
Diffusion Model (DMs)~\cite{ddpm2020neurips,song2020score} are constructed through a two-phase process: a {forward diffusion process} and a {backward denoising process}. In the forward diffusion process, DDPM~\cite{ho2020denoising} transforms a clean image $\mbx_0$ into a noisy image $\mbx_t$ at each time step $t$ as:
\begin{align}
    \mbx_t := \alpha_t \mbx_0 + \sigma_t \mbepsilon, \mbepsilon\sim \cN(\mbepsilon; \mbzero, \mbI)
\end{align}
where $\alpha_t$ is the diffusion coefficient that controls the scaling of the data, and $\sigma_t^2$ is the variance of noise at time step $t$. We can take $\sigma_t=\sqrt{1-\alpha_t}$ for variance-preserving diffusion model, such as DDPM~\cite{ho2020denoising}. This forward diffusion process is designed to gradually corrupt the image, transforming it into a distribution that is approximately Gaussian as $t$ approaches to $1$. 

DMs' goal is to learn a neural network $\mbepsilon_{\mbtheta}(\mbx_t, t)$ that approximates the score function of the noise-corrupted data, $\nabla_{\mbx_t} \log p_{\text{real}}(\mbx_t)$, where $p_{\text{real}}(\mbx_t)$ is the probability density of the noisy data at timestep $t$, in which the score function can be written as
$\nabla_{\mbx_t} \log p_{\text{real}}(\mbx_t) = -\sigma_t^{-1}(\mbx_t - \alpha_t \mbx_0)$. 

Training DMs is executed by minimizing a weighted Root Mean Square Error~(RMSE):
\begin{align}
\E_{t, \mbx_t}\left[\omega(t)\|\mbepsilon_{\mbtheta}(\alpha_t \mbx_0 + \sqrt{1-\alpha_t} \mbepsilon, t) - \mbepsilon\|^2\right],    
\end{align}
where $\omega(t)$ is a weighting function that balances the contribution of different time steps to the overall loss.

Once the model is trained, data generation is achieved through an iterative denoising process. Starting from $\mbx_1\sim\cN(\mbx_1;\mbzero, \mbI)$, DMs uses $\mbepsilon_{\mbtheta}(\mbx_t, t)$ predicts the noise component and then update the noisy image. This process continues until $t$ reaches $0$, at which point the generated image is expected to follow the distribution of clean images. The iterative nature of this process, while effective in producing high-quality images, is computationally expensive, as it requires multiple evaluations of the neural network. 

\subsubsection{Denoising Diffusion Implicit Model~(DDIM)} In addition to to the backward denoising process mentioned above, DDIM~\cite{song2020denoising} provides deterministic sampling procedures to accelerate the denoising process by defining the denoising step as:
\begin{multline} \label{eq:ddim-sampling}
    \mbx_{t-1} = \sqrt{{\alpha}_{t-1}} \mbx_t - \frac{\sqrt{1 - {\alpha}_t} \, \epsilon_{\theta}(\mbx_t, t)}{\sqrt{{\alpha}_t}} \\
    + \sqrt{1 - {\alpha}_{t-1}} \, \epsilon_{\theta}(\mbx_t, t)   
\end{multline} 
DDIM enables a faster sampling procedure with
a small impact on the quality of the generated samples


\begin{figure*}[ht]
    \includegraphics[trim=65 145 40 85, clip, width=\linewidth]{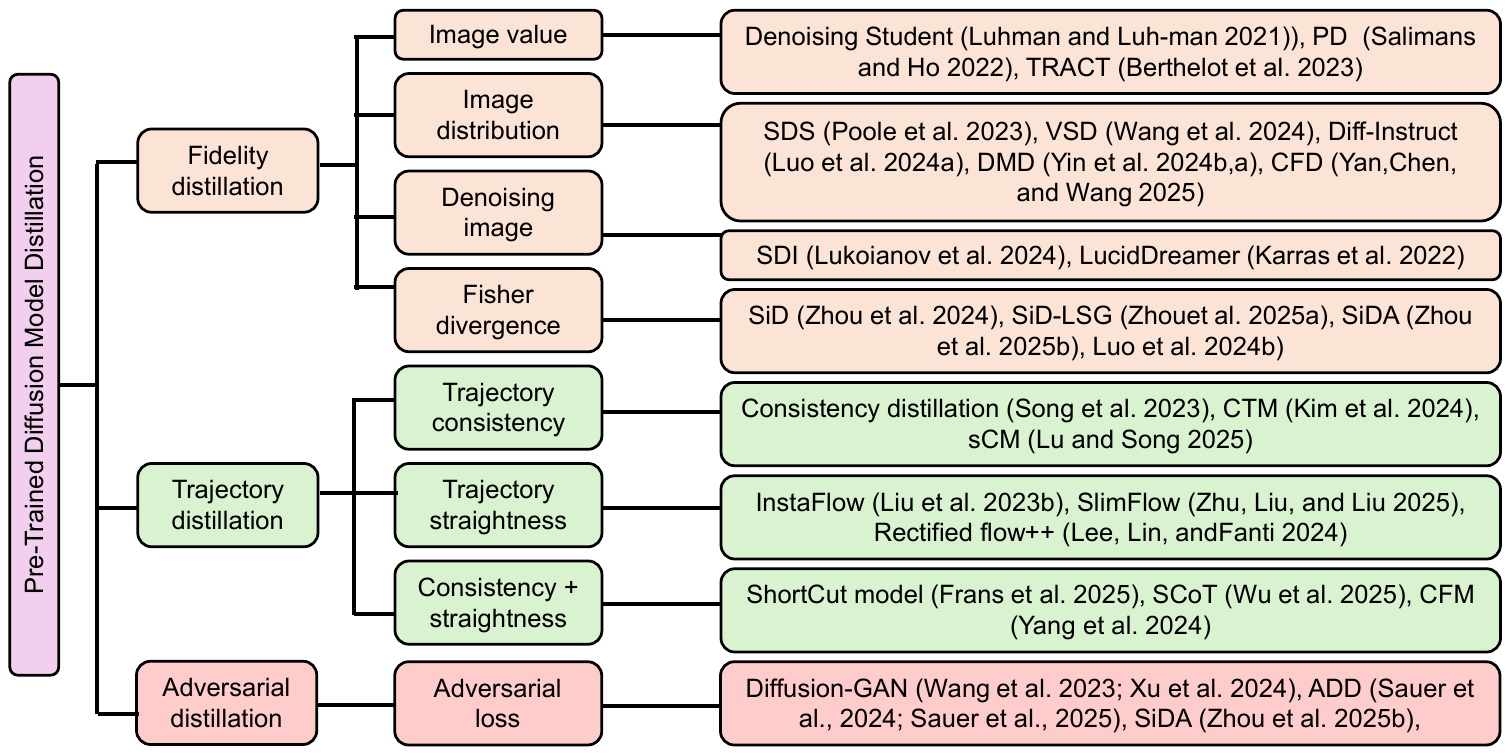}
    \caption{Taxonomy of Pre-trained Diffusion model distillation methods.}
    \centering
    \label{fig:taxonomy}
\end{figure*}

\section{Overview of Pre-trained DMs Distillation}
\label{sec:general-distillation}
Given a pre-trained DMs as the teacher model, distillation aims to obtain a student model that is typically more compact in size and can use fewer steps to generate high-quality images. We categorize existing pre-trained DMs distillation methods based on their motivations: fidelity distillation, trajectory distillation, and adversarial distillation.

Existing work on fidelity distillation tries to minimize the differences between outputs from the teacher model and from the student model. These output differences can be measured by different criteria such as RMSE measuring the two models' output image values, Kullback-Leibler divergence~(KL-diverence) measuring two models' image distributions, difference between the gradient of log-likelihood for the images (i.e. Fisher divergence). 

The trajectory distillation methods monitor the entire trajectory between random noises $\mbx_1$ and clean images $\mbx_0$. It targets at optimizing the pre-trained trajectory for fast image sampling. Based on the trajectory properties they want to achieve, these methods can be categorized into consistency distillation, rectified flow distillation and methods that integrate both properties. The first focuses on using self-consistency property to regularize trajectory, while the second enforces the trajectory to be straight for fast sampling.

DMs is firstly used as a generator in the adversarial loss, in which an alternative optmization strategy is used by interleaves optimizing the classifier's parameter and the DM's parameter. It is then used as an independent distillation method. Adversarial loss can be used as a plug-and-play tool in other distillation methods to further improve their performance. 

Figure~\ref{fig:taxonomy} summarizes the general categories of these distillation methods. 

\section{Fidelity Distillation}
\label{sec:output-distillation}
Fidelity distillation aims to maintain the quality and faithfulness of the generated output. This is achieved by minimizing the discrepancy between the outputs of the teacher model and the student model.

\subsection{Output Reconstruction Loss}
\subsubsection{Denoising student} Denoising student~\cite{luhman2021knowledge} may be the first to distill pre-trained DMs. It proposes a one-step generator and optimizes the generator by minimizing the RMSE between the outputs from the one-step generator and the teacher model. Although promising, the distilled one-step generator might generate low-quality images due to the one step restriction.  

\subsubsection{Progressive Distillation} Progressive Distillation~\cite{salimans2022progressive,meng2023distillation} recursively halves the number of sampling steps. During each halving step, the student model is optimized such that its one-step projection closely approximates the to teacher model's two-step projection. 

\subsubsection{TRACT} Building on Progressive Distillation, TRACT~\cite{berthelot2023tract} first divides the time interval $[0, 1]$ into several segments and then trains a student model to distill the output of a teacher model’s inference output from step $t_i$ to $t_j (t_j<t_i)$ within each segment. By repeating this procedure, both progressive distillation and TRACT can learn a one-step generator. 

\subsection{Output Distribution Loss}
In output distribution loss, a one-step generator $f(\mbphi)$ is designed to represent the clean image as $\mbx_0=f(\mbphi)$. Optimizing $\mbphi$ can be achieved by involving the teacher model to calculate the gradient of the loss with respect to $\mbphi$. 

\subsubsection{Score Distillation Sampling~(SDS)} SDS~\cite{poole2022dreamfusion} is initially introduced to work on the text-to-$3$D image synthesis task. By replacing the clean image $\mbx_0$ with the one-step generator $f(\mbphi)$, the loss function can be written as:
$\cL_{\text{SDS}}=\E_{t, {\mbepsilon}}\left[\omega(t)\|\mbepsilon_{\mbtheta}(\alpha_tf(\mbphi)+\sigma_t\mbepsilon, t) - \mbepsilon\|^2\right]$.     
Omitting the Jacobian of $\mbepsilon_{\mbtheta}(\mbx_t, t)$, i.e. $\partial \mbepsilon_{\mbtheta}(\mbx_t, t)/\partial \mbx_t$, the gradient of $\cL_{\text{SDS}}$ w.r.t. $\mbphi$ is:
\begin{align} \label{eq:sds-update-form}
\nabla_{\mbphi}{\cL_{\text{SDS}}}:=\E_{t, {\mbepsilon}}\left[\omega(t)\left(\mbepsilon_{\mbtheta}(\alpha_t\mbx_0+\sigma_t\mbepsilon, t) - \mbepsilon\right)\frac{\partial \mbx_0}{\partial \mbphi}\right]  \end{align}
Built on SDS, Consistent Flow Distillation~(CFD)~\cite{yan2025consistent} introduces multi-view consistent Gaussian noise on the 3D object.

An important explanation of SDS is that its gradient $\nabla_{\mbphi}{\cL_{\text{SDS}}}$ may be viewed as a KL-divergence between the proposal distributions and ground-truth distributions $\nabla_{\mbphi}{\cL_{\text{SDS}}} \propto \E_t\left[\KL{q(\mbx_t|f(\mbphi), t)}{p_{\mbtheta}(\mbx_t|t)}\right]$. This insight has inspired the following output distribution loss work. 

\subsubsection{Variational Score Distillation~(VSD)} Instead of targeting at the point estimator for $\mbphi$, VSD~\cite{wang2024prolificdreamer} targets at optimizing $\mbphi$'s distribution. The gradient can be written as
$\nabla_{\mbphi}{\cL_{\text{VSD}}}\approx\E_{t, \mbx_t}\left[\omega(t)(\mbepsilon_{\mbtheta}(\alpha_t\mbx_0+\sigma_t\mbepsilon, t) - \mbepsilon_{\mbpsi}(\mbx_t, t))\frac{\partial \mbx_0}{\partial \mbphi}\right]$y, where $\mbepsilon_{\mbpsi}(\mbx_t, t)$ is another noise prediction network, where its parameter $\mbpsi$ can usually initialized by the teacher model. 

Training VSD involves an alternative optimization strategy which interleaves optimizing the generator's parameter $\mbphi$ and the noisy score's parameter $\mbpsi$. It is noted that SDS is a special case of VSD when we specify $\mbphi$'s distribution as a Dirac distribution $\delta(\mbphi-\mbphi^*)$. Comparing to SDS, VSD introduces an extra term $\mbepsilon_{\mbpsi}(\mbx_t, t)$, $\mbepsilon_{\mbpsi}(\mbx_t, t)$ is able to provide more modelling capability, such as text embedding input, than one single Gaussian noise $\mbepsilon$.

\subsubsection{Diff-Instruct} Diff-Instruct~\cite{luo2024diff} is concurrently developed with VSD. Diff-Instruct develops an integrated KL-divergence as the objective function to optimize the one-step generator: $\cL_{\text{DI}}=\int_{t=0}^1\omega(t)\E_{\mbx_t\sim q^t(\mbx_t)}\left[\log \frac{q^{t}(\mbx_t)}{p^{t}(\mbx_t)}\right]\mathrm{d}t$, where $q^t(\mbx_t)$  denotes the marginal densities of the diffusion process at time step $t$, initialized with standard Gaussian distributions, and $p^t(\mbx_t)$ denotes the data distribution. Training Diff-Instruct works similar to VSD, which also interleaves optimizing the generator's parameter $\mbphi$ and an additional noisy score's parameter $\mbpsi$. Score implicit Matching~(SiM)~\cite{luo2024one} further develops Diff-Instruct with a time-integral score-divergence. 

\subsubsection{Distribution Matching Distillation~(DMD)} The KL-divergence in SDS coincides with DMD~\cite{yin2024one,yin2024improved}, which also considers the KL-divergence between the synthetic data generator $p_{\text{fake}}(\mbx_t)$ and the real data distribution $p_{\text{real}}(\mbx_t)$ as $\cL_{\text{DMD}}=\KL{p_{\text{fake}}(\cdot)}{p_{\text{real}}(\cdot)}=\E_{\mbz, \mbx, t, \mbx_t}\left[\log \frac{p_{\text{fake}}(\mbx_t)}{p_{\text{real}}(\mbx_t)}\right]$, where $\mbz\sim\cN(\mbzero, \mbI), \mbx_0 = g_{\mbphi}(\mbz), t\sim\text{Unif}(0, 1), \mbx_t=\alpha_t\mbx_0+\sigma_t\mbepsilon$. $\cL_{\text{DMD}}$'s gradient with respect to $\mbphi$ can be conveniently calculated as $\nabla_{\mbphi}\KL{p_{\text{fake}}(\cdot)}{p_{\text{real}}(\cdot)}=\E_{\mbx_t}\left[(\mbs_{\text{fake}}(\mbx_t)-\mbs_{\text{real}}(\mbx_t))\frac{\partial \mbx_0}{\partial \mbphi}\right]$. 

Building on DMD, \cite{nguyen2024swiftbrush} further demonstrates its effectiveness in distilling pre-trained stable DMs and extends its application beyond 3D synthesis. Meanwhile, \cite{xie2024distillation} introduces a maximum likelihood-based approach for distilling pre-trained DMs, where the generator parameters are updated using samples drawn from the joint distribution of the diffusion teacher prior and the inferred generator latents.

\subsection{One-step Denoising Image Space}
\subsubsection{Score Distillation Inversion~(SDI)} Using SDS may lead to the so-called over-smoothing phenomenon~\cite{liang2024luciddreamer}, due to the high-variance random noise term $\mbepsilon$ and also the large reconstruction error. SDI~\cite{lukoianov2024score,liang2024luciddreamer} is proposed to work in the one-step denoising image space $\mbx_0(t)= \mbx_t - \sigma_t\epsilon_{\mbtheta}(\mbx_t, t)$, where $\mbx_0(t)$ is the noisy data at time step $t$ denoised with one single step of noise prediction.

Applying $\mbx_0(t)$ into the guidance term $(\mbepsilon_{\mbtheta}(\alpha_t\mbx+\sigma_t\mbepsilon, t) - \mbepsilon)$ of Equation~(\ref{eq:sds-update-form}), the updated guidance term $(\mbepsilon_{\mbtheta}(\alpha_t\mbx_0+\sigma_t\mbepsilon, t) - \mbepsilon)$ can be formulated as $\mbx_0(t-\tau)=\mbx_0(t)-\sigma_t\left[\mbepsilon_{\mbtheta}(\mbx_{t-\tau}, t-\tau)-\mbepsilon_{\mbtheta}(\mbx_{t}, t)\right]$. 
After a series of transformations, the gradient with respect to $\mbphi$ can be written as $\nabla_{\mbphi}{\cL_{\text{SDI}}}:=\E_{t, \mbx_t}\left[\omega(t)\left(\mbepsilon_{\mbtheta}(\alpha_t\mbx_0+\sigma_t\mbepsilon, t) - \mbkappa_t\right)\frac{\partial \mbx_0}{\partial \mbphi}\right]$, where $\mbkappa_t:=\text{DDIM-Inversion}(\mbx_t,t)$ can be obtained through DDIM inversion.

\subsubsection{Diffusion Distillation based on Bootstrapping~(Boot)} Boot~\cite{gu2023boot} also works on the one-step denoising image space. While initially inspired from the consistency model~\cite{song2023consistency} which seeks a deterministic mapping from any noisy image $\mbx_t$ to clean image $\mbx_0$, Boot does the opposite by mapping any noisy image $\mbx_t$ from the a random noise $\mbx_1$. Training Boot on $\mbx_0(t)$ helps to focus on the low-frequency ``signal'' of the noisy image $\mbx_t$ and stablise the training process. Of course, Boot may also be regarded as an extension from consistency model~(See panel (c) in Figure~\ref{fig:visualisation-on-trajectory-distillation-methods}). 

\subsection{Fisher Divergence Loss}
\subsubsection{Score identity Distillation~(SiD)} SiD~\cite{zhou2024score} uses a model-based score-matching loss as the objection function, which moves away from the traditional usage of KL-divergence between real and fake distributions over noisy images. This model-based loss (Fisher divergence) can be written in an $L2$-loss as $\cL_\theta(\phi^*, \psi^*(\theta), t) 
= \mathbb{E}_{\mbx_t \sim \rho_{p_\theta}(\mbx_t)} \left[ \frac{\alpha_t^2}{\sigma_t^2} \| f_{\phi^*}(\mbx_t, t) - f_{\psi^*(\theta)}(\mbx_t, t) \|_2^2 \right]$, where $f_{\phi^*}(\mbx_t, t)$ and $f_{\psi^*}(\mbx_t, t)$
represent the optimal denoising score networks for the true and fake data
distributions, respectively. The SiD approache is claimed to be data-free, and has shown impressive empirical performance, as it has demonstrated the potential to
match or even surpass the performance of teacher models in a single generation step. 

Subsequent work based on SiD including SiD-LSG~\cite{zhou2024long} which develops a long and short classifier-free guidance strategy to distill pre-trained models using only synthetic images generated by the one-step generator; Score identity distillation with adversarial~(SiDA)~\cite{zhou2024adversarial} which further incorporates adversarial training strategy to improve the model performance; Score implit Matching~(SiM)~\cite{luo2024one} extends the $L2$-loss into more general format, by developing a score-gradient theorem to calculate the gradient of those more general divergences. 

\begin{figure*}
    \centering
    \includegraphics[width=1\linewidth]{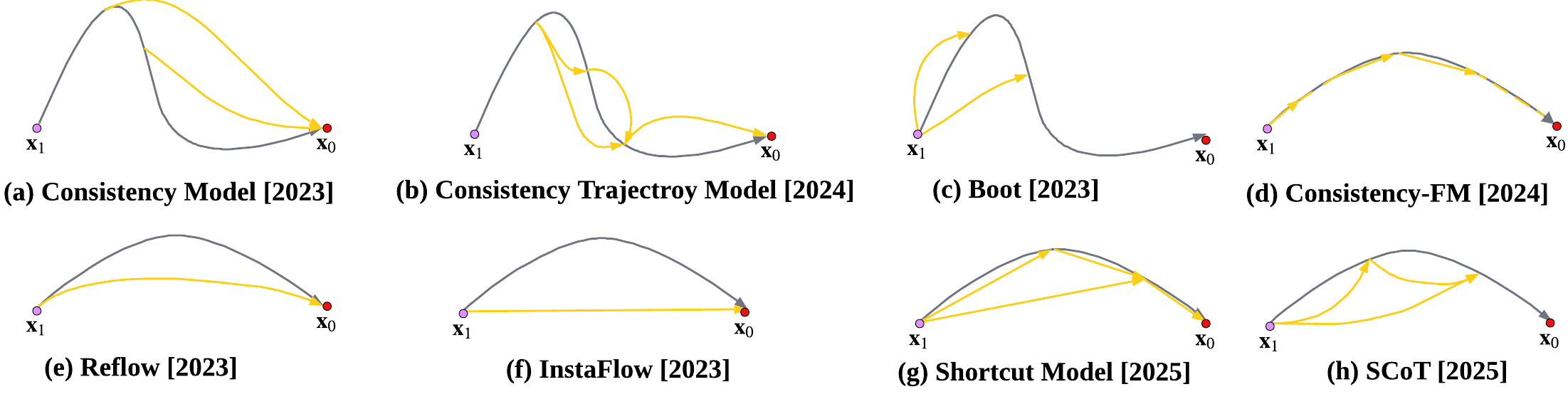}
    \caption{Visualisations on trajectory distillation methods. Grey arrows represent the trajectories of pre-trained models, whereas yellow arrows correspond to the trajectories of the distillation models. Panels (a-c) illustrate consistency distillation methods, while panels (d-h) depict rectified flow distillation methods. The trajectories in (a-c) exhibit greater curvature compared to those in (d-h), reflecting the fact that the pre-trained models in (a-c) are DMs, while those in (d-h) are rectified flows. In panels (b), (g), and (h), a single large step corresponds to the position of two smaller steps due to the self-consistency property of the trajectory. }
    \label{fig:visualisation-on-trajectory-distillation-methods}
\end{figure*}

\section{Trajectory Distillation}
\label{sec:trajectory-distillation}
Instead of generative fidelity, trajectory distillation focuses on the mapping trajectory between the noise distribution and the clean image distribution. Based on different requirements and properties for trajectories, we discuss two types: consistency distillation for trajectory self-consistency, and rectified flow distillation for straight trajectory. Figure~\ref{fig:visualisation-on-trajectory-distillation-methods} demonstrates a visualization of these consistency distillation and rectified flow distillation methods. 

\subsection{Consistency Distillation}
\subsubsection{Consistency model~(CM)} CM~\cite{song2023consistency} is developed based on the probability flow-ODE~\cite{song2020score}. CM learns a consistency function $f_{\mbphi}(\mbx_t, t)$ which maps an noisy image $\mbx_t$ back to the clean image $\mbx_0$ as $f_{\mbphi}(\mbx_t, t) = \mbx_0$. When the consistency function $f_{\mbphi}(\mbx_t, t)$ is required to satisfy the boundary condition at time step $t=0$, $f_{\mbphi}(\mbx_t, t)$ is usually parameterized as $f_{\mbphi}(\mbx_t, t) = c_{\text{skip}}(t)\mbx_t+c_{\text{out}}(t)F_{\mbphi}(\mbx_t, t)$, where $F_{\mbphi}$ is the actual neural network to train, and $c_{\text{skip}}(t)$ and $c_{\text{out}}(t)$ are time-dependent factors such that $c_{\text{skip}}(0) = 1, c_{\text{out}}(0) = 0$. 

Although CM can be trained from scratch, a more popular choice for CM is to do distillation on the trajectory from teacher model as consistency distillation~(CD). In particular, the distillation objective function can be written as a distance metric between adjacent points as $\cL_{\text{CD}} = \E_{i}\left[\omega(t_i)d(\mbf_{\mbphi}(\mbx_{t_{i}+\Delta t}, t_{i}+\Delta t), \mbf_{\mbphi^-}(\widehat{\mbx}^{\mbtheta}_{t_{i}}, t_{i}))\right]$, 
where $d(\cdot, \cdot)$ is a metric function, $\mbphi^-$ is the exponential moving average (EMA) of the past values $\mbphi$, and $\widehat{\mbx}^{\mbtheta}_{t_{i}}$ is obtained from pre-trained model as
$\widehat{\mbx}^{\mbtheta}_{t_{i}}=\mbx_{t_i}-(t_i-t_{i+1})t_{i+1}\nabla_{\mbx_{t_{i+1}}}\log p_{t_{i+1}}(\mbx_{t_{i+1}})$. 

As consistency models directly map noisy inputs to their clean counterparts, the iterative denoising process is avoided and it may generate high-quality images with fewer steps.  

\subsubsection{Consistency Trajectory Model~(CTM)} CTM~\cite{kim2023consistency} is introduced to minimize the accumulated estimation errors and discretization inaccuracies in multi-step consistency model sampling. While the consistency function in CM projects noisy image $\mbx_t$ at any time step $t$ to its clean image $\mbx_0$, CTM extends it by allowing the projection function to any earlier time step $s (s<t)$ as $f_{\mbphi}(\mbx_t, t, s)$. In order to satisfies the boundary condition when $f_{\mbphi}(\mbx_t, t, 1) = \mbx_1$, $f_{\mbphi}(\mbx_t, t, 1)$ is parameterized as $ f_{\mbphi}(\mbx_t, t, s)= \frac{s}{t}\cdot\mbx_t+\frac{1-s}{t}\cdot F_{\mbphi}(\mbx_t, t, s)$, where $F_{\mbphi}()$ is the actual neural networks to be optimized.

Regarding loss function, CTM minimizes a {soft consistency matching loss}, which is defined as $\cL_{\text{CTM}} = \|f_{\mbphi}(\mbx_t, t, s)-f_{\text{sg}(\mbphi)}(f_{\mbtheta}(\mbx_t, t, u), u, s)\|^2$, where $0<s<u<t<1$, and $\text{sg}(\mbphi)$ is the EMA stop gradient operator. In $\cL_{\text{CTM}}$, the one projected from the teacher model and the one from the student model should projected to the same value at the same time step. 

\subsubsection{Continuous-time Consistency Model} Although most consistency models (CM) use discretized time steps for training, this approach often suffers from issues of tuning additional hyperparameters and inherently introduces discretization errors. \cite{song2023improved,geng2024consistency,lu2024simplifying} have been working on continuous-time CM and strategies to stablize its training. For example, \cite{geng2024consistency} analyses CM from a differential perspective and then progressively tightens the consistency condition, by letting the time different $\Delta t\to \mathrm{d}t$ as training progresses. \cite{lu2024simplifying} proposes a TrigFlow framework, which tries to unify the flow matching~\cite{lipman2022flow} and DMs, by defining the noisy image as $\mbx_t=\cos(t)\mbx_0+\sin(t)\mbepsilon$ for $t\in[0, \pi/2]$ and setting the training objective as $\cL_{\text{TrigFlow}}=\E_{\mbx_0, \mbz, t}\left[\|\sigma_t F_{\mbphi}(\mbx_t/\sigma_t, t)-\mbx_t\|^2\right]$, where $F_{\mbphi}(\cdot)$ is the actual neural networks to be optimized. 

\subsection{Rectified Flow}
\subsubsection{Rectified Flow} Rectified Flow~\cite{lipman2022flow,liu2023flow,albergo2023stochastic}, also named as flow matching or stochastic interpolation, leverage Ordinary Differential Equations~(ODEs) to model the transition between two distributions, i.e. $p_0(\cdot), p_1(\cdot)$. Given a clean image $\mbx_0 \sim p_0(\mbx_0)$ and a random noise $\mbx_1 \sim p_1(\mbx_1)$, rectified flows construct a linear interpolation path defined as $\mbx_t = t \mbx_1 + (1 - t) \mbx_0, \quad 0 \leq t \leq 1$. 

The training objective for the model is formulated as:
\begin{align} \label{eq:rectified-flow-first-training}
    \cL_{\text{rf}}(\mbtheta)= \int \mathbb{E}_{\mbx_1,\mbx_0}\left[\|(\mbx_1 - \mbx_0) - \mbv_{\mbtheta}(\mbx_r, r)\|^2\right] dr,
\end{align}
where $\mbv_{\mbtheta}(\mbx_t, t)$ represents the learned and pre-trained velocity field. 

Given a random noise $\mbx_1$, rectified flow's new image generation process follows the ODE as $d\mbx_t = \mbv_{\mbtheta}(\mbx_t, t)\mathrm{d}t, t\in [0, 1]$, where we can write the recovered clean image $\widehat{\mbx}_0$ as $\widehat{\mbx}_0=\mbx_1 + \int_{1}^0\mbv_{\mbtheta}(\mbx_r, r)\mathrm{d}r$.  
Through minimizing $\cL_{\text{rf}}$, the trained velocity field $\mbv_{\mbtheta}(\mbx_t, t)$ is expected to approximate the direct path from $\mbx_1$ to $\mbx_0$. That is, rectified flows may be able to achieve high-quality data generation with fewer steps compared to traditional DMs, where the trajectory is usually curved and need more time steps to generate images. Further studies include MeanFlow~\cite{geng2025mean} which defines an average velocity $\mbu(\mbz_t, s, t) := \frac{1}{t-r}\int_{r}^t\mbv(\mbz_{r}, r)\mathrm{d}r$ to describe the velocity field, rather than the instantaneous velocity from rectified flows, and rectified flow++~\cite{lee2024improving} further improves rectified flow by developing a U-shaped timestep distribution and LPIPS-Huber premetric. 

\subsubsection{Reflow vs InstaFlow} In fact, while $\mbx_1$ and $\mbx_0$ are independently sampled from $p_0(\mbx_0)$ and $p_1(\mbx_1)$, simply training Equation~(\ref{eq:rectified-flow-first-training}) may not be able to obtain a straight trajectory. This is because the random pair $(\mbx_1, \mbx_0)$ may provide highly noisy or even contradictory information regarding the trajectory between $\mbx_1$ and $\mbx_0$. Thus, \cite{liu2023flow} proposes a reflow mechanism, which first construct a data pair $(\mbx_1, \widehat{\mbx}_0)$ from the pre-trained model $\mbv_{\mbtheta}(\mbx_t, t)$,  where $\widehat{\mbx}_0$ is defined above as $\widehat{\mbx}_0=\mbx_1 + \int_{1}^0\mbv_{\mbtheta}(\mbx_r, r)dr$, and then trains a new velocity $\mbv_{\mbphi}(\mbx_t, t)$ as $\cL_{\text{reflow}} = \int \mathbb{E}_{\mbx_1}\left[\|(\mbx_1 - \widehat{\mbx}_0) - \mbv^{1}_{\mbphi}(\mbx_t, t)\|^2\right] \mathrm{d}t$, where $\mbx_t = (1-t)\widehat{\mbx}_0+t\mbx_1$. 

On the other hand, InstaFlow~\cite{liu2023instaflow,zhu2025slimflow} aims to construct a one-step generator as $\mbx_1 + \mbv_{\mbphi}(\mbx_1, 1)$, where $\mbv_{\mbphi}(\mbx_1, 1)$ can be trained as $    \cL_{\text{rf-dis}} =   \mathbb{E}_{\mbx_1}\left[\|(\mbx_1 - \widehat{\mbx}_0) - \mbv^{1}_{\mbphi}(\mbx_1, 1)\|^2\right]$. 

In fact, we can see that $\mbv_{\mbphi}(\mbx_1, 1)$ is learned to approximate the amount of velocity $\mbv_{\mbtheta}(\mbx_r, r)$ changes from $t=1$ to $t=0$. Comparing these two distillation methods, reflow tries to obtain a new trajectory, where each time step's velocity is close to constant, whereas Instaflow tries to approximate the amount of changes given the random noise $\mbx_1$.  

\subsection{Integrating Consistency and Straightness}
Consistency models and rectified flows each have distinct advantages and limitations. The former excels in regulating trajectory behavior, while the latter prioritizes achieving a straight trajectory.  

\subsubsection{Shortcut model} Shortcut model~\cite{frans2025one} is recently proposed as a promising improvement over the rectified flow. Shortcut model made two critical modifications to the existing rectified flow: (1) in addition to considering the current noisy image $\mbx_t$ and its corresponding time step $t$, Shortcut model also takes the future projected time step $t+d$ as an input for the neural network $\mbv_{\mbphi}(\mbx_t, t, d)$. As a result, this future projected time step can help guide the generation of future denoising images; (2) Shortcut model also regulates the trajectory to be self-consistent in the velocity perspective. Its loss function composes of a reflow distillation $\E_{\mbx_1, t, d}[\|(\mbx_1 - \widehat{\mbx}_0) - \mbv_{\mbphi}(\mbx_t, t, 0)\|^2$ and a self-consistency loss $\|\mbv_{\mbphi}(\mbx_t, t, 2d)-\frac{\mbv_{\mbphi}(\mbx_t, t, d)+\mbv_{\mbphi}(\mbx'_{t+d}, t, d)}{2}\|^2]$, where $\mbx'_{t+d} = \mbx_t + \mbv_{\mbphi}(\mbx_t, t, d)$. The former term enforces constant velocities, while the latter term ensures the self-consistency of the generated trajectory. 

\subsubsection{Straight Consistent Trajectory~(SCoT)} SCoT~\cite{wu2025traflow} also targets at trajectory distillation to produce ones that is straight and consistent at the same time. Instead of approximating velocities in Shortcut model, SCoT chooses to approximate the trajectory itself by adopting CTM's projection function. SCoT discovers the connection between CTM and rectified flows by observing that $\frac{\partial f_{\mbphi}(\mbx_t, t, s)}{\partial s} = \mbv_{\mbphi}(\mbx_s)$. The loss function of TraFlow considers the factors of soft consistency matching and trajectory straightness. 
Since SCoT directly projects the noisy image at a future time step $s$, it removes the need for an ODE solver to approximate the trajectory, thereby avoiding approximation errors.

In fact, both Shortcut model and SCoT share the same motivation and even model structure with CTM~\cite{kim2023consistency}, which takes the noisy image $\mbx_t$, the current time step $t$, and the future time step $s$ or the step lag $s-t$ as inputs. Integrating the consistency model with trajectory straightness is a natural choice, as a straight trajectory is inherently self-consistent.

\subsubsection{Consistency-FM~(CFM)} Instead of requiring the trajectory to be self-consistent, Consistency-FM~\cite{yang2024consistency} works on the velocity side by setting consistent velocities. In particular, consistency-FM derives the equivalence between two conditions at any two time steps $\forall t,s\in[0, 1]$: 
\begin{align}
    \text{condition 1: } &\mbv(\mbx_t, t) =\mbv(\mbx_s, s)\nonumber; \\
    \text{condition 2: } &\mbx_t+(1-t)\mbv(\mbx_t, t)= \mbx_s+(1-s)\mbv(\mbx_s, s), \nonumber
\end{align}
in which condition 1 means the velocities at any time steps are the same and condition 2 refers to that the linearly projected points from different time steps would be the same. Consistency-FM set the training objective as to let the velocity satisfies these two conditions. Consistency-FM also develops multi-segment mechanism to further improve the model performance.

\section{Adversarial Loss}
\label{sec:adversarial-distillation}
Diffusion-GAN~\cite{wang2022diffusion,xu2024ufogen} may be the first introducing the adversarial loss~\cite{goodfellow2020generative} in training DMs. Diffusion-GAN optimizes a min-max objective function to obtain an optmized DM:
\begin{multline}
    \cL_{\text{Diffusion-GAN}} = \E_{t, \mbx_t\sim q(\mbx|\mbx_0, t)}\left[\log \mathcal{D}_{\mbpsi}(\mbx_t)\right] \\
    + \E_{t, \mbz\sim p(\mbz), \mbx'_t\sim q(\mbx'_t|f_{\mbphi}(\mbz), t)}\left[\log(1-\mathcal{D}_{\mbpsi}(\mbx'_t))\right],
\end{multline}
where $\mathcal{D}_{\mbpsi}(\cdot)$ is the classifier for adversarial loss. The training process involves alternating optimization of the generator's parameters, $\mbphi$, and the classifier's parameters, $\mbpsi$. 

In addition to direct training, adversarial loss itself can be independently used as adversarial distribution distillation~(ADD)~\cite{sauer2025adversarial}, in which the loss function is in the GAN format:
\begin{multline}
    \cL_{\text{ADD}} = \E_{t, \mbz\sim p(\mbz), \mbx'_t\sim q(\mbx'_t|f_{\mbphi}(\mbz), t)}\left[\log \mathcal{D}_{\mbpsi}(\mbx_t)\right. \\
    \left. + \log(1-\mathcal{D}_{\mbpsi}(\mbx'_t))\right].
\end{multline}
ADD has been successfully used in SDXL Turbo~\cite{sauer2024fast}, a text-to-image model that enables real-time generation. 

Since this adversarial loss is orthogonal to existing methods, common distillation methods can include it as an additional technique to improve the distillation performance~\cite{zhou2024adversarial}. When real data is available, the performance of the student model might surpass that of the teacher model.

\section{Challenges and Future Direction}
\label{sec:challenges-future-directions}
Although pre-trained DMs distillation methods have made significant progress, this field still faces substantial challenges, along with promising future directions. 

\subsection{Challenges}
\subsubsection{1, Training on large pre-trained models} Most experiments evaluating the effectiveness of distillation methods are conducted on small datasets, such as CIFAR-10 and ImageNet. However, distilling large-scale pre-trained DMs, such as Stable Diffusion~\cite{rombach2022high,esser2024scaling}, remains computationally demanding and is infeasible for research groups with limited GPU access. Therefore, developing computationally efficient approaches for large-scale model distillation is a critical challenge.

\subsubsection{2, Trade-off between quality and speed} Distillation methods accelerate sampling by reducing the number of function evaluations~(NFE) or sampling steps. As a result, The reduction in NFE might lead to degraded image quality due to the accummulated errors, as it is commonly to see that multi-step generation works better than one-step generation even in distillation methods. Obtaining an optimal balance requires carefully designed approximations, improved training objectives, and architectural modifications. 

\subsubsection{3, Training guidelines} The lack of uniform training guidelines for DMs distillation creates inconsistencies in methodology and evaluation. While it is common to adopt the settings of EDM~\cite{karras2022elucidating,karras2024analyzing} to train DMs, key distillation aspects, such as weight functions, network architectures, and distillation stages, vary across studies without a standardized framework. Different weight functions alter training dynamics, while architectural choices and stage configurations affect efficiency and fidelity.

\subsection{Future directions}

\subsubsection{1, Training smaller-size student model} 
Most existing approaches primarily focus on reducing the number of time steps to achieve high-quality image generation. However, they often overlook effective strategies for reducing model size, specifically the neural network’s capacity. Employing a smaller student model is particularly beneficial for practical applications in low-resource environments. SlimFlow~\cite{zhu2025slimflow} explores this direction by using annealing strategy to reduce model size. Further such exploration is encouraged.

\subsubsection{2, Trajectory optimization} Optimizing the trajectory between random noise and clean images remains a crucial area, particularly with the advancement of consistency models and rectified flows. Current approaches primarily focus on enforcing self-consistency and trajectory straightness. However, further exploration of alternative trajectory properties could provide valuable insights for enhancing both the efficiency and quality of sampling.

\subsubsection{3, Student model initialization} The parameters of the student model are typically initialized using those of the teacher model. However, the two models are designed with different objectives and sampling mechanisms. For instance, a student model with only a few sampling steps can be viewed as an integration of a teacher model with many sampling steps. While effective initialization can accelerate student model training, the development of a universal initialization strategy remains an open research question.

\subsubsection{4, Theoretical understanding of using distillation}
There is a question of whether distillation is necessary to achieve such models or if they could be trained directly. While the existence of a distilled model demonstrates feasibility, it does not ensure optimal efficiency in achieving it. Some models obtained through diffusion distillation can, in principle, be trained from scratch, though this approach generally performs worse when compared to distillation, reflecting the well-established advantage of distilling large neural networks into smaller ones. This analogy holds for DMs, where distillation compresses a deep sampling process into a shallower one. Consistency models might offer an example where both distillation and direct training are possible, but the latter requires careful hyperparameter tuning and training strategies. In addition to such empirical evidence, a theoretical understanding of the advantages of distillation methods would be interesting to explore in the future.

\subsubsection{5, Broader applications of distillation methods} This survey primarily focuses on the methodological aspects of distillation techniques and does not address practical applications such as text-to-image synthesis. Given the demonstrated effectiveness of DMs in fields like image, audio, and video generation, it would be exciting to explore further applications in these domains and beyond. Particularly, the unique advantages of distillation methods in adapting to low computational resource environments present significant potential for broader application. 

\clearpage
\newpage
\bibliography{aaai2026}

\end{document}

%% file: math.tex
\newcommand{\mathbold}[1]{\ensuremath{\boldsymbol{\mathbf{#1}}}}

\newcommand{\nestedmathbold}[1]{{\mathbold{#1}}}

\newcommand{\mbf}{\nestedmathbold{f}}

\newcommand{\mbs}{\nestedmathbold{s}}

\newcommand{\mbu}{\nestedmathbold{u}}
\newcommand{\mbv}{\nestedmathbold{v}}

\newcommand{\mbx}{\nestedmathbold{x}}

\newcommand{\mbz}{\nestedmathbold{z}}

\newcommand{\mbI}{\nestedmathbold{I}}

\newcommand{\mbepsilon}{\nestedmathbold{\epsilon}}

\newcommand{\mbkappa}{\nestedmathbold{\kappa}}

\newcommand{\mbphi}{\nestedmathbold{\phi}}

\newcommand{\mbpsi}{\nestedmathbold{\psi}}

\newcommand{\mbtheta}{\nestedmathbold{\theta}}

\newcommand{\mbzero}{\nestedmathbold{0}}

\DeclareRobustCommand{\KL}[2]{\ensuremath{\textsc{kl}\left[#1\;\|\;#2\right]}}

\newcommand{\cL}{\mathcal{L}}
\newcommand{\cN}{\mathcal{N}}

\newcommand{\E}{\mathbb{E}}

